\documentclass[conference]{IEEEtran}
\IEEEoverridecommandlockouts
\usepackage{hyperref}
\usepackage{graphicx}
\usepackage{subfigure}
\usepackage{setspace}
\usepackage{xcolor}
\usepackage{multicol}
\DeclareUnicodeCharacter{2212}{-}
\DeclareUnicodeCharacter{2009}{-}
\usepackage{amsmath}
\usepackage{amssymb}
\usepackage{float}

\usepackage{caption}
\usepackage{cite}
\usepackage{amsmath,amssymb,amsfonts}
\usepackage{graphicx}
\usepackage{array}
\usepackage{booktabs}
\usepackage{caption}
\usepackage{ragged2e}
\usepackage{makecell} 
\usepackage{adjustbox} 
\usepackage{anyfontsize} 
\usepackage{booktabs}
\usepackage{array}
\usepackage{tabularx} 
\usepackage{array}
\usepackage{graphicx}
\usepackage{textcomp}
\usepackage{xcolor}
\usepackage{multirow}
\usepackage{booktabs}
\usepackage{graphicx}
\usepackage{multirow}
\usepackage{threeparttable}

\usepackage{xcolor}
\usepackage{cite}
\usepackage{amsmath,amssymb,amsfonts}
\usepackage{algorithmic}
\usepackage{graphicx}
\usepackage{textcomp}
\usepackage{xcolor}
\def\BibTeX{{\rm B\kern-.05em{\sc i\kern-.025em b}\kern-.08em
    T\kern-.1667em\lower.7ex\hbox{E}\kern-.125emX}}
    
\begin{document}

\title{Empowering Meta-Analysis: Leveraging Large Language Models for Scientific Synthesis}
\author{\IEEEauthorblockN{Jawad Ibn Ahad}
\IEEEauthorblockA{\textit{Apurba-NSU R\&D Lab, ECE} \\
\textit{North South University}\\
Dhaka, Bangladesh \\
jawad.ibn@northsouth.edu}
\and
\IEEEauthorblockN{Rafeed Mohammad Sultan}
\IEEEauthorblockA{\textit{Apurba-NSU R\&D Lab, ECE} \\
\textit{North South University}\\
Dhaka, Bangladesh \\
rafeed.sultan@northsouth.edu}
\and
\IEEEauthorblockN{Abraham Kaikobad}
\IEEEauthorblockA{\textit{Apurba-NSU R\&D Lab, ECE} \\
\textit{North South University}\\
Dhaka, Bangladesh \\
abraham.kaikobad@northsouth.edu}
\and
\IEEEauthorblockN{Fuad Rahman}
\IEEEauthorblockA{\textit{Apurba Technologies}\\
Sunnyvale, CA 94085, USA \\
fuad@apurbatech.com}
\and 
\IEEEauthorblockN{Mohammad Ruhul Amin}
\IEEEauthorblockA{\textit{Computer and Information Science} \\
\textit{Fordham University}\\
New York, USA \\
mamin17@fordham.edu}
\and
\IEEEauthorblockN{Nabeel Mohammed}
\IEEEauthorblockA{\textit{Apurba-NSU R\&D Lab, ECE} \\
\textit{North South University}\\
Dhaka, Bangladesh \\
nabeel.mohammed@northsouth.edu}
\and 
\IEEEauthorblockN{Shafin Rahman}
\IEEEauthorblockA{\textit{Apurba-NSU R\&D Lab, ECE} \\
\textit{North South University}\\
Dhaka, Bangladesh \\
shafin.rahman@northsouth.edu}
}

\maketitle

\begin{abstract}
This study investigates the automation of meta-analysis in scientific documents using large language models (LLMs). Meta-analysis is a robust statistical method that synthesizes the findings of multiple studies (\emph{support articles}) to provide a comprehensive understanding. We know that a \emph{meta-article} provides a structured analysis of several articles. However, conducting meta-analysis by hand is labor-intensive, time-consuming, and susceptible to human error, highlighting the need for automated pipelines to streamline the process. Our research introduces a novel approach that fine-tunes the LLM on extensive scientific datasets to address challenges in big data handling and structured data extraction. We automate and optimize the meta-analysis process by integrating Retrieval Augmented Generation (RAG). Tailored through prompt engineering and a new loss metric, Inverse Cosine Distance (ICD), designed for fine-tuning on large contextual datasets, LLMs efficiently generate structured meta-analysis content. Human evaluation then assesses relevance and provides information on model performance in key metrics. This research demonstrates that fine-tuned models outperform non-fine-tuned models, with fine-tuned LLMs generating 87.6\% relevant meta-analysis abstracts. The relevance of the context, based on human evaluation, shows a reduction in irrelevancy from 4.56\% to 1.9\%. These experiments were conducted in a low-resource environment, highlighting the study's contribution to enhancing the efficiency and reliability of meta-analysis automation.

\end{abstract}


\section{Introduction}
Meta-analysis is a powerful statistical approach that combines
the findings of multiple studies to provide a comprehensive
understanding of the same research topic~\cite{Jadotte_Moyer_Gurevitch_2023}. A meta-analysis paper, or \emph{\textbf{meta article}}, offers a structured analysis of numerous individual \emph{\textbf{support articles}}. Individual studies often face limitations, such as small sample sizes or narrow focus, making it hard to draw definitive conclusions~\cite{article}. Meta-analysis aggregates data from different studies, providing robust estimates that inform research decisions, guide treatments, and influence healthcare policies~\cite{moreau2022conducting, phillippo2019threshold, wiles2022consumer, gopalakrishnan2013systematic}. In applied scientific fields, meta-analyses play a crucial role: consolidating the results of clinical trials~\cite{haidich2010meta, lee2018overview}, evaluating public health strategies, material performance, and farming practices~\cite{talic2021effectiveness, toledo2019effect, lee2019impact}, and assessing the impacts, behavior, policies, and teaching methods of climate in fields such as environmental science, psychology, economics, and education~\cite{osbaldiston2012environmental, van2021systematic, wehkamp2018governance, wisniewski2020power}. Meta-analysis involves the analysis of extensive datasets, as it incorporates numerous studies, which presents a significant challenge to big data. The process is often labor-intensive, requiring manual extraction and analysis of data from multiple research articles, which is both time-consuming and susceptible to human error. This underscores the critical need for an automated pipeline to streamline and improve the efficiency of meta-analysis generation. The advancements in large language models (LLMs) offer promising potential, suggesting that these models could be utilized to manage the vast data requirements of scientific meta-analysis.

\begin{figure*}[!t]
    \centering
    \includegraphics[scale=0.47]{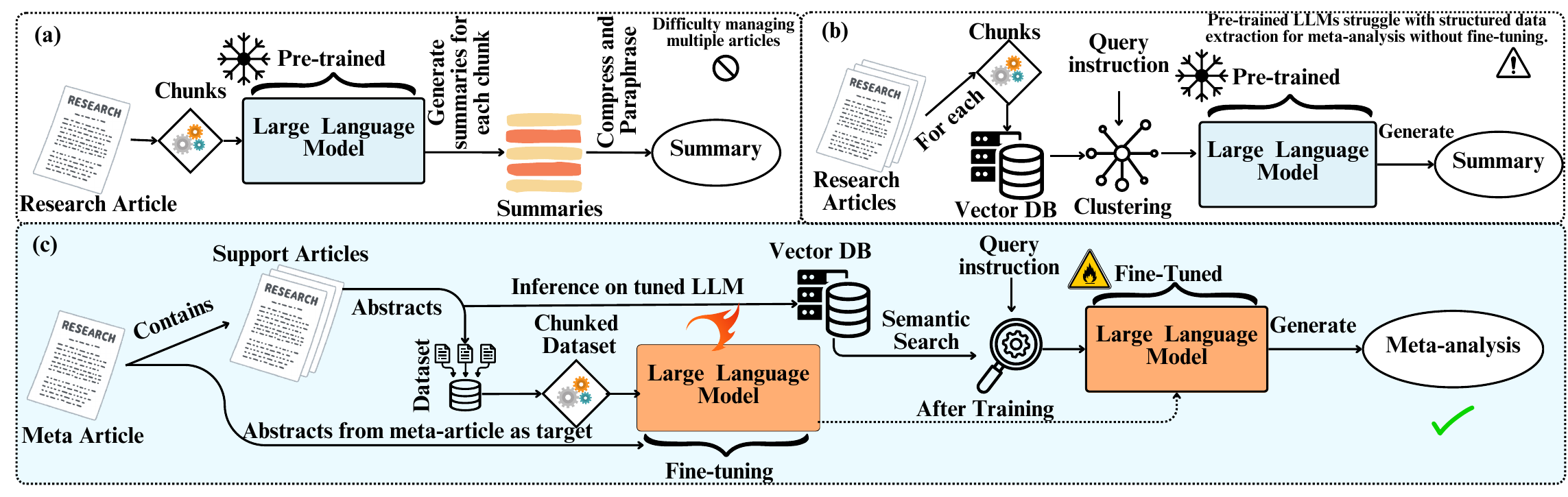}
    \caption{(a) Paraphraser-based approach that combines multiple generated summary chunks from LLMs has been used by \cite{Subbiah2024ReadingSE, lim2023improving}, (b) Retrieval augmentation generation-based approach has been applied in \cite{yepes2024financial, manathunga2023retrieval} using a vector database to store chunked data and cluster them before passing to LLM to produce a summary. Existing methods often fall short of handling big scientific contextual data and generating structured synthesis. (c) We propose a novel approach involving fine-tuning LLMs with large contexts and utilizing them to generate meta-analysis abstracts. Abstracts from \textit{support papers} serve as input, with \textit{meta-papers'} abstracts as labels. Pre-processing involves chunking the dataset due to context length restrictions and prioritizing small LLMs over resource-intensive large LLMs. The fine-tuned model generates meta-analysis abstracts via semantic search with the provided context and query.}
    \label{fig:hybrid}
\end{figure*}

Traditional meta-analysis faces challenges with big data scalability. \emph{\textbf{(a) Manual Meta-analysis:}} Involves manual data extraction and analysis, limiting scalability in rapidly expanding domains~\cite{chen2022intervention}. 
While AI advances automate tasks like information retrieval and summarization, focusing primarily on shorter summaries, LLMs show strong potential in summarization but still face limitations. \emph{\textbf{(b) LLMs as summarizers:}} 
 Although LLMs like Llama2-13b have demonstrated proficiency in generating summaries and performing question-answering tasks~\cite{keswani2024abstractive, Subbiah2024ReadingSE, lim2023improving}, their utility in synthesizing extensive research findings for meta-analysis remains constrained. Current approaches in this domain predominantly focus on generating condensed summaries for shorter, narrative-based content rather than synthesizing large-scale scientific data. To address this gap, researchers have explored RAG techniques. \emph{\textbf{(c) Retrieval Augmented Generation:}} By integrating retrieval-based mechanisms, RAG enables LLMs to access and summarize large datasets through document retrieval~\cite{Wei2023EvaluationOC, yepes2024financial, manathunga2023retrieval}. However, this method falls short when applied to meta-analysis, which demands specialized data extraction techniques and a deeper synthesis of scientific contexts~\cite{Reason2024ArtificialIT}. LLMs' current limitations highlight a need for targeted fine-tuning and tailored approaches to handle complex, structured large-scale scientific data. 

To bridge this gap, our research introduces a novel approach that leverages LLMs with RAG to automate and streamline the meta-analysis process. We have built a comprehensive dataset with various meta-analysis scenarios in various scientific fields, which contains the content of the meta-articles along with the content of the support papers. This dataset facilitates both training and evaluation to stimulate further research. Its purpose is to fine-tune LLMs, enabling them to understand and replicate data extraction patterns for meta-analysis. We introduce a novel loss function, \emph{Inverse Cosine Distance (ICD)}, specifically designed for training LLMs in large-context scenarios to handle large-data challenges. This function enhances the performance of LLMs in generating meta-analysis with high relevance and accuracy. By fine-tuning LLMs for large-context tasks and employing specific prompt engineering techniques, shown in Fig.~\ref{fig:hybrid}, we aim to overcome the limitations of existing methods of handling big contextual data. By integrating RAG with our fine-tuning strategy, LLMs generate precise, instruction-based meta-analysis content, ensuring quality and efficiency. Our approach reduces the labor-intensive aspects of meta-analysis, enabling LLMs to handle large contexts and generate structured abstracts effectively. This work holds significant potential for improving research synthesis across various domains.

Our contribution comprises \emph{\textbf{(1)}} preparing a comprehensive dataset to fine-tune LLMs for meta-analysis generation, \emph{\textbf{(2)}} fine-tuning LLMs with the novel ICD loss function, enhancing their ability to handle large-context scientific data and extract relevant information for meta-analysis, and \emph{\textbf{(3)}} leveraging these fine-tuned LLMs by integrating RAG to generate precise, instruction-based meta-analysis from large-scale scientific data.

\section{Related works}
\noindent \textbf{Meta-Analysis Strategy: }
In recent years, the landscape of meta-analysis has witnessed significant advancements, particularly with the development of comprehensive databases, facilitating systematic reviews. Csizmadia et al. \cite{Csizmadia2022AGD} contributed to this domain by introducing a global database of innovation and quality management, meticulously compiling the PRISMA methodology, covering records from 1975 to 2021. Furthermore, Yudhanto et al. \cite{yudhanto2021metadata} addressed the labor-intensive nature of data collection for meta-analysis, presenting a method using bibliometric studies from the Science Direct Database. Their approach, which involved data collection from published searches using desired keywords over the last decade, significantly streamlined the process, contributing to the efficiency of meta-analysis procedures. To further enhance this progress, our paper introduces a novel approach that harnesses the capabilities of LLMs and  RAG. This approach aims to streamline the meta-analysis process, empower LLMs to handle large contexts efficiently, and conduct a structured meta-analysis of the provided research papers. Our contribution lies in developing a Comprehensive Meta-Analysis Dataset, which serves as a valuable resource for training and evaluating the efficiency of our proposed approach.

\noindent \textbf{Large Context Summarization:}
Recent advancements in large-context summarization have led to the development of techniques to generate concise and informative summaries from extensive documents. In particular, Subbiah et al. \cite{Subbiah2024ReadingSE} introduce a fragmentation strategy, converting full stories into manageable fragments and associating them with prompts to facilitate effective summarization. Keswani et al. \cite{keswani2024abstractive} focus on summarization and question-answering tasks using the Llama-2 (13B) model, employing clustering techniques based on cosine similarity to improve efficiency despite computational constraints. Furthermore, observations on different LLM summarization performances have been made using zero-shot prompt techniques \cite{Goyal2022NewsSA, Zhang2023BenchmarkingLL}. These advancements underscore the significance of leveraging LLMs for large-context summarization tasks. Leveraging these developments, our proposed method successfully utilizes large contexts by breaking them into smaller, manageable chunks. This strategy facilitates easier summarization by smaller LLMs, thereby enhancing the overall efficiency of the process.

\noindent \textbf{Summarization Quality Assessment by Evaluation:} Evaluation metrics are crucial for assessing the effectiveness of automated summarization systems. Traditional metrics often have limitations in accurately capturing the quality of generated summaries~\cite{Ermakova2019ASO}. Innovative frameworks like HumanELY \cite{Reprint2023HUMANELYHE} have been proposed, incorporating key evaluation metrics including relevance, coverage, coherence, harm, and comparison. Additionally, novel scoring systems leveraging LLMs have been introduced, shedding light on how different identities influence performance \cite{Lu2023CharacterisedLA}. A taxonomy of LLM-based NLG evaluation methods has also been presented, delineating their advantages and drawbacks \cite{Gao2024LLMbasedNE}. Despite these efforts, achieving comprehensive evaluation frameworks for NLG systems remains challenging. Inspired by the pioneering work of Chaudhary et al. \cite{Chaudhary2023ItsAR} on generating both relevant and irrelevant queries, we adopt a similar methodology to evaluate the efficacy of our generated meta-analysis. Our evaluation metrics encompass not only relevance but also nuances, categorizing outputs into \textit{Relevant}, \textit{Somewhat-Relevant}, and \textit{Irrelevant}. Incorporating this thorough evaluation framework allows us to deliver a detailed evaluation of the performance of our automated meta-analysis synthesis, thereby enriching the depth of analysis and insights in our research. Our approach introduces a novel evaluation concept based on hard voting, contributing to meta-analysis automation. Leveraging a comprehensive meta-analysis dataset, innovative training methods, and fine-tuning strategies tailored for instruction-based meta-analysis abstracts, our approach stands out for its effectiveness and reliability in advancing research synthesis efficiency across domains.

\section{Methodology}

Several innovative efforts have been made to guarantee that LLMs can manage lengthy contexts. However, to incorporate big textual data challenges, LLMs require numerous amounts of resources.  
This study presents a novel approach for generating meta-analysis using LLMs, particularly with long context lengths. This section formally outlines our method for using LLM to produce meta-analysis content.\\
\noindent\textbf{Problem Formulation:}
Consider there are $m^j$ number of meta-articles, where $j \in [1, n]$. For each meta-article, there is a set of support articles, $S^j = = \{v^j_1, v^j_2,...,v^j_{|S^j|}\}$. $v^j_i$ represents the abstract of $i^{th}$ support article related to $j^{th}$ meta-article.
We aim to build a model $\mathcal{M}$ to 
generate a meta-article's abstract, $y^j$ using all abstracts inside the set $S^j$. This study focuses on generating a relevant $y^j$ through a low context length LLM (e.g., Llama-2 7B, Mistral 7B, and Gemma 7B).

Here we have investigated two important aspects of this problem. \textbf{(a) \emph{Handling large context length: }} Typically, meta-analyses are performed through manual analysis and data extraction from supporting articles. Recently, LLMs have demonstrated their ability to summarize extensive textual data. While substantial research has focused on summarization~\cite{Wei2023EvaluationOC, yepes2024financial, manathunga2023retrieval}, the application of LLMs for meta-analysis remains unexplored. Meta-analysis often involves structured data derived from supporting articles, yet most LLMs operate within constrained context-length environments during fine-tuning. Our objective is to address this limitation by efficiently managing large contextual data and segmenting it into smaller chunks to facilitate effective fine-tuning of LLMs. \textbf{(b) \emph{Enhance information retrieval: }}Prior research has demonstrated that fine-tuning LLMs enhances their data extraction capabilities~\cite{Reason2024ArtificialIT}. However, when generating large contextual, analytical data, LLMs require access to external knowledge sources. RAG has shown promising results in addressing this challenge. In our approach to enabling context-length-restricted LLMs to generate meta-analysis and to further expand the scope of knowledge from supporting articles, we aim to integrate RAG with our fine-tuned meta-analysis generator LLMs.

\begin{figure*}[t!]
\includegraphics[width=\textwidth]{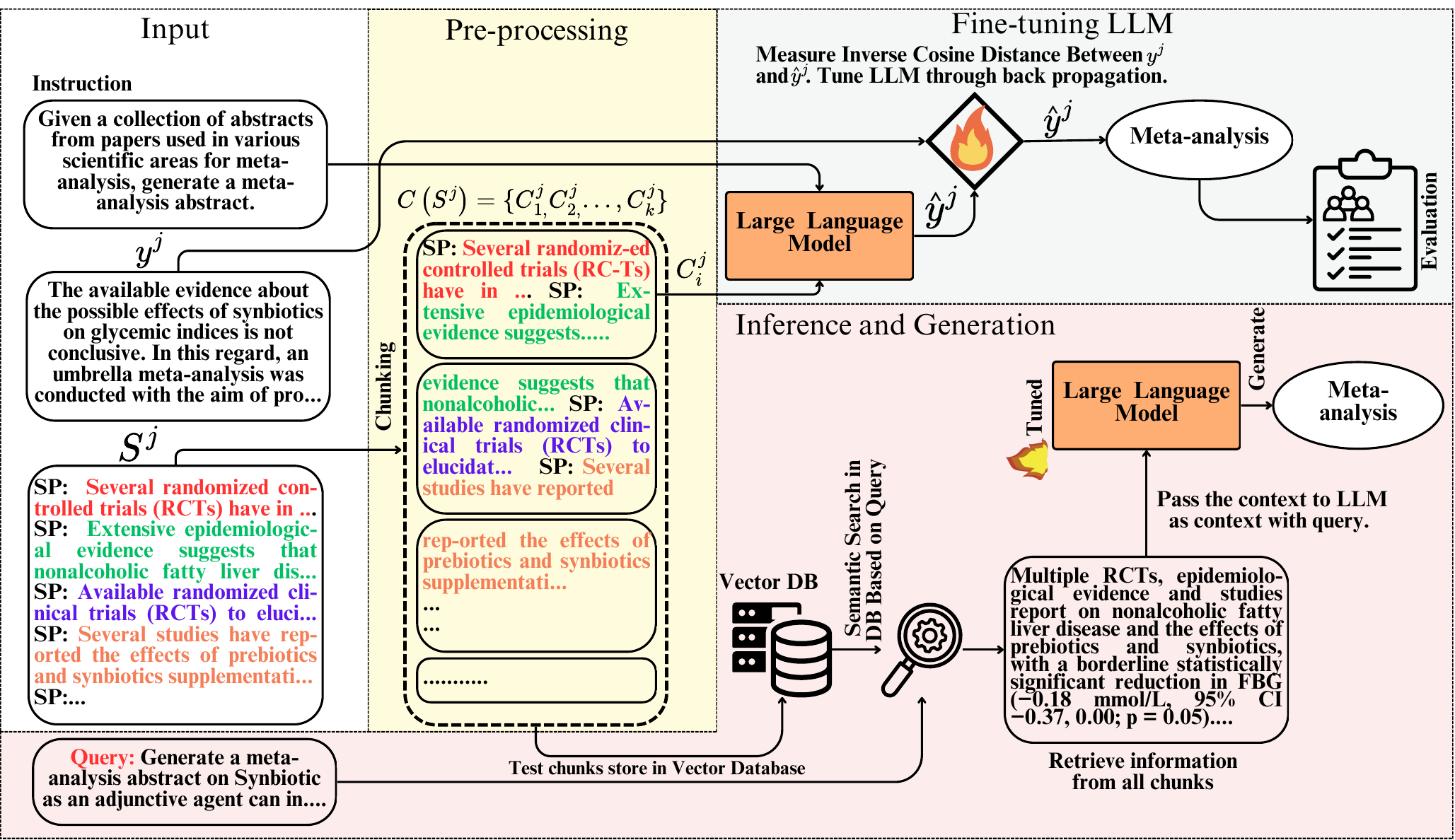}
\caption{In our meta-analysis generation system, support articles \( S^j \) undergo chunk-based pre-processing, producing chunks \( C^j_i \subseteq S^j \), here ``SP:" refers to an abstract of the support article \( S^j \). These chunks are used to fine-tune the LLMs for predicting meta-analysis abstracts \( y^j \), with the ICD loss guiding the fine-tuning process. Model performance is assessed through human evaluation of the relevancy of generated meta-analysis abstracts, $\hat{y}^j$ by fine-tuned LLMs. During inference, we integrate RAG with the fine-tuned LLMs. Chunked samples are stored in a vector database, from which relevant information is retrieved via a semantic search based on a query. The same processed \( C^j_i\) is used for both fine-tuning and inference to maintain retrieval consistency. The retrieved content and the query are provided to the LLM, enabling it to generate more precise and accurate meta-analysis abstracts by leveraging comprehensive contextual information.}
\label{archi}
\end{figure*}
\subsection{MAD: Meta-Analysis Dataset}
Generative language models are capable of producing reviews or summaries of given contexts. They still have issues producing analytical context based on substantial context inputs. Our first step in addressing this big-data challenge is to create a dataset to generate a meta-analysis. Large scientific context datasets like \textbf{MAD} have not been used before to fine-tune context-length-restricted LLMs.

The dataset, \textbf{MAD} that we constructed consists of two columns: one containing \textit{meta-articles'} abstracts and the other containing the abstracts of the \textit{support articles}. For example, consider the meta-article titled ``Intervention methods for improving reduced heart rate variability in patients with major depressive disorder: A systematic review and meta-analysis"~\cite{chen2022intervention}. We used the abstract of this paper as our target meta-analysis abstract. From \emph{Table 1} of this paper, we identified that it conducted a meta-analysis of over twenty studies. We manually extracted the abstracts of these support articles by following the references listed in the table. These twenty abstracts were placed in the second column ($S^j$) alongside the meta-article's abstract ($y^j$). Essentially, the goal is for the LLM to generate a meta-analysis abstract from these support articles' abstracts. 

\begin{figure}[!t]
    \centering
    \includegraphics[trim={1.8cm 0.2cm 0.0cm 1cm},scale = 0.38, clip]{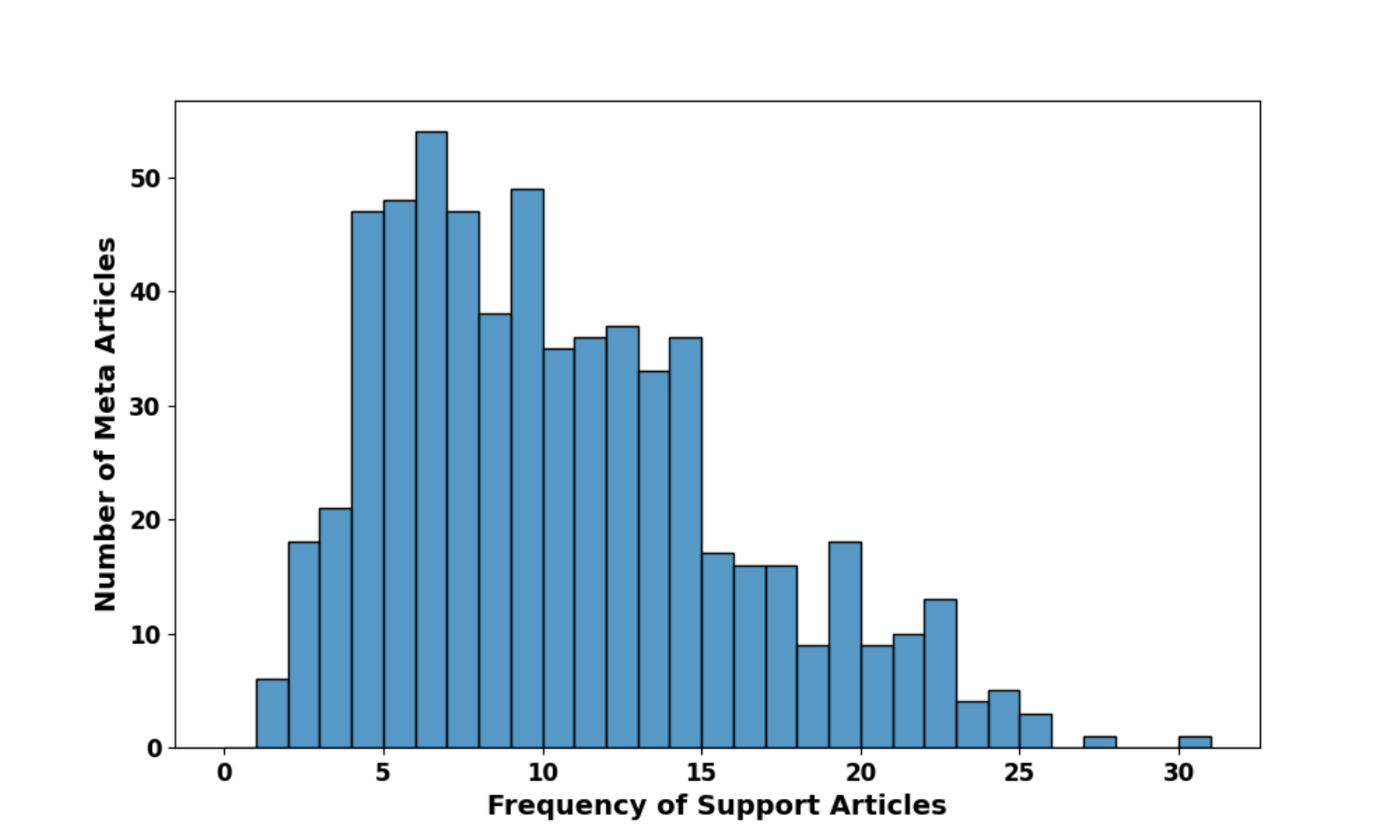}
    \caption{Distribution of Supporting Articles in Meta-Articles in the dataset \textbf{MAD}. The chart shows that most meta-articles contain  6 to 14 support articles, with peaks at 6 and 9, suggesting a common reliance on a moderate number of supporting studies, with fewer analyses incorporating larger study pools.}
    \label{fig:freqsupport}
\end{figure}

Using this approach, we gathered 625 meta-articles from ScienceDirect, along with the abstracts of all the support articles included in that meta-analysis.  
In total, dataset \textbf{MAD} includes 6344 support articles' abstracts and 625 meta-articles' abstracts. The dataset statistics, along with the demographic information of the human evaluators who assessed model performance, are shown in Table \ref{tab:stat}, and the distribution of support articles in meta-articles is shown in Fig.~\ref{fig:freqsupport}.
\subsection{Chunk-Based Processing of Support Articles}
Given the limitation in context length for many language models, processing long or complex documents as a whole can become inefficient and may lead to suboptimal results. To address this, chunking the support articles into smaller, meaningful segments allows for more effective input to the language model. 
By chunking, we ensure that all support article abstracts in the set 
$S^j$ are considered while maintaining manageable input sizes for low-context models. 

To manage the input size and improve model performance, we divide the support articles set \( S^j = \{v^j_1, v^j_2, \dots, v^j_{|S^j|} \} \) into multiple smaller overlapping chunks. Overlapping will be done with some portions of abstracts. This will allow the coherence and continuity between chunks, reducing the chances of information loss.  
Chunking of \( S^j \) into \( k \) possibly overlapping chunks is defined as:
   \[
   \mathcal{C}(S^j) = \{ C^j_1, C^j_2, \dots, C^j_k \}
   \]
   where:
   \begin{itemize}
       \item \( C^j_i \subseteq S^j \) for each \( i \in [1, k] \),
        \item  \( \bigcup_{i=1}^{k} C^j_i = S^j \) (the union of all chunks covers the entire set, though the chunks may overlap),
   \item \( C^j_i \cap C^j_l \neq \emptyset \) for \( i \neq l \) (contents will overlap). 
   \end{itemize}
For example, suppose the set of support article abstracts \( S^j = \{v^j_1, v^j_2, v^j_3, v^j_4, v^j_5 \} \) is divided into three overlapping chunks. The chunking is as follows: $\mathcal{C}(S^j) = \{ C^j_1, C^j_2, C^j_3 \}$ where:$C^j_1 = \{v^j_1, v^j_2\}, \quad C^j_2 = \{(portion\, of)v^j_2, v^j_3, v^j_4\}, \quad C^j_3 = \{{(part\,of)}v^j_4, v^j_5\}.$
In this example, \( v^j_2 \) and \( v^j_4 \) overlap in \( C^j_2 \) and \( C^j_3 \). 

\subsection{Fine-tune LLMs and Integrate RAG}
\begin{table}[!t]
\centering
\caption{Detailed statistics of the actual and processed dataset \textbf{MAD}, along with the demographic profile of Human Evaluators for assessing the readability of fine-tuned LLMs. The \textbf{MAD} dataset contains abstracts from medical meta-articles and support studies. }
\label{tab:stat}
\setlength{\tabcolsep}{0.8em} 
\renewcommand{\arraystretch}{0.75} 
\scalebox{0.85}{ 
\begin{tabular}{c}
\toprule
\textbf{MAD Statistics} \\
\begin{tabular}{@{}cc@{}}
\multicolumn{2}{c}{\begin{tabular}{@{}c|c|c@{}}
 & \textbf{Actual} & \textbf{Chunked} \\ \hline
Min. input $(S^j)$ context length & 733  & 1005 \\
Max. input $(S^j)$ context length & 32767 & 2000 \\
Avg. input $(S^j)$ context length & 16890.22 & 1542.32 \\
Min. labels $(y^j)$ context length & 104 & 104 \\
Max. labels $(y^j)$ context length & 2492 & 2492 \\
Avg. labels $(y^j)$ context length & 1446.45 & 1446.45 \\
Total Instances & 625 & 7447 \\
\end{tabular}} \\
\end{tabular} \\
\midrule[\heavyrulewidth]
\textbf{Human Evaluators Details} \\
\midrule
\begin{tabular}{@{}cc@{}}
\textbf{Total no. of evaluators} & 13 \\
\textbf{No. of female evaluators} & 4 \\
\textbf{No. of male evaluators} & 9 \\
\textbf{Avg. age} & 23 \\
\textbf{Profession} & Student, Engineer \\
\textbf{Education Background} & Undergraduate \\
\end{tabular} \\
\bottomrule
\end{tabular}}
\end{table}

After creating the dataset MAD, two popular LLMs are considered for the experiment: Llama-2 (7B)\cite{Touvron2023Llama2O} and Mistral-v0.1 (7B)\cite{Jiang2023Mistral7}. They are fine-tuned on the constructed dataset MAD and evaluated using the test set. To further improve their performance in generating relevant outcomes, we applied the RAG approach~\cite{Lewis2020RetrievalAugmentedGF} to the fine-tuned versions of each model. Fig~\ref{archi} depicts the methodology for fine-tuning LLMs and generating meta-analysis.

\noindent\textbf{Model Architecture Overview:} We utilized two prominent LLMs in this study. \textbf{(a)} Llama-2 (7B), a transformer-based LLM developed by Meta, includes 32 attention heads, a 32,000-token vocabulary, and a context length of 4,096. It uses the Swish-Gated Linear Unit (SwiGLU) activation function~\cite{Shazeer2020GLUVI}. \textbf{(b)} Mistral-v0.1 (7B) features similar architecture with 32 attention heads and a 32,000-token vocabulary but offers a larger context length of 8,192. It employs the Sigmoid Linear Unit (SiLU) activation function~\cite{Elfwing2017SigmoidWeightedLU} and incorporates grouped-query attention (GQA) with sliding window attention (SWA) for efficient handling of variable sequences~\cite{Jiang2023Mistral7}. Mistral-v0.1 (7B) outperforms both Llama-2 (7B) and Llama-2 (13B) in benchmarks, making it our model of choice.

\noindent\textbf{Fine-tuning LLMs:}
Original models like Llama-2 and Mistral-v0.1 benefit from extensive pre-training on massive datasets, allowing them to grasp complex linguistic structures. However, this generic training might not be ideal for specialized tasks like generating meta-analysis abstracts from lengthy source materials. Fine-tuning bridges this gap by adapting these pre-trained models to new datasets and large data tasks.

Following the selection of models, Llama-2 (7B) and Mistral-v0.1 (7B) were fine-tuned on the processed MAD dataset, utilizing chunked samples $C^j_i$ paired with their respective meta-article's abstracts $y^j$. This supervised fine-tuning process paired each chunked sample with its meta-analysis abstract $(C^j_i,y^j)$, where \( C^j_i \subseteq S^j \) and $y^j$ serves as label meta-article's abstract. The models, $\mathcal{M}$, were trained to recognize patterns for generating meta-analysis content from large contexts, accommodating the multiple chunks associated with each $y^j$. Instruction-based fine-tuning was employed with a focus on prompt engineering. Various prompt configurations were tested to optimize the models' ability to generate accurate and coherent meta-analysis abstracts from long contexts. This approach ensured that the models effectively learned the specific patterns required for high-quality content generation.

\noindent\textbf{Inverse Cosine Distance (ICD):}\label{customloss} To support fine-tuning, a specialized training mechanism is used with the ICD loss metric. The ICD function measures the dissimilarity between the model-generated output $\hat{y}$ and the ground truth $y$ vectors, incorporating a small positive constant $\epsilon$ in the denominator to ensure numerical stability and improve the fine-tuning process.
The formulation for ICD is given below:
\begin{equation} \label{icd}
     \text{ICD} = \frac{1}{N} \sum_{i=1}^{N} \frac{1}{cosine\_sim_i(y,\hat{y})+\epsilon}
\end{equation}

During fine-tuning, models $\mathcal{M}$ processed chunked samples \( C^j_i \) to produce predicted abstracts \( \hat{y}^j \). The ICD loss, calculated using formula~\ref{icd}, measured the dissimilarity between \( \hat{y}^j \) and the ground truth abstracts \( y^j \), guiding parameter updates via backpropagation. The process, constrained by resources, was carried out for 2 epochs over 5 iterations, refining the model and improving abstract accuracy.

\begin{table*}[!t]
\begin{threeparttable}[b]
\centering
\small
\caption{After fine-tuning the LLMs on the dataset \textbf{MAD}, comparing model performance on benchmark datasets for summarization quality, enabling assessment across varying context lengths. Among all the limited context-length LLMs, our fine-tuned (FT) models performed acceptably well comparatively. On the scientific document-contained dataset, Cl-SciSumm, our fine-tuned model outperformed other pre-trained LLMs, showing a significantly rigorous capability of capturing structured analytical information. $\uparrow$ ($\downarrow$) means higher (lower) is better. `-' denotes results that are not applicable there.}\label{tab1}
\renewcommand{\arraystretch}{0.8}
\setlength{\tabcolsep}{1.1em}

\begin{tabular}{@{}cccccccc@{}}
\toprule
 \textbf{Method} &\textbf{Models} & \multicolumn{2}{c}{\textbf{Open-i}\tnote{1}} & \multicolumn{2}{c}{\textbf{writer\_summaries}\tnote{2}} & \multicolumn{2}{c}{\textbf{CL-SciSumm}\tnote{3}} \\
\cmidrule(lr){3-4} \cmidrule(lr){5-6} \cmidrule(lr){7-8}
 & & BLEU $\uparrow$ & ROUGE $\uparrow$ & BLEU $\uparrow$ & ROUGE $\uparrow$ & BLEU $\uparrow$ & ROUGE $\uparrow$ \\
\midrule
\multirow{3}{*}{Established\tnote{4}} & GPT-4 with ICL~\cite{Veen2023AdaptedLL} & 46.0 & 68.2 & -- & -- & -- & -- \\
& InstructGPT davinci v2~\cite{Zhang2023BenchmarkingLL} & -- & -- & -- & \>48 & -- & -- \\
& GCN Hybrid~\cite{yasunaga2019scisummnet} & -- & -- & -- & -- & -- & 33.88 \\
\midrule
\midrule
\multicolumn{8} {c} {\bfseries Context length restricted LLMs} \\ \midrule
Pre-trained & Falcon 7B~\cite{almazrouei2023falcon} & 0.19 & 3.17 & 0.76 & 5.19 & 0.71 & 2.21 \\
Pre-trained & Gemma 7B~\cite{team2024gemma} & 2.13 & 8.81 & 4.47 & 30.28 & 2.44 & 20.78 \\
Pre-trained & Orca-2 7B~\cite{mitra2023orca} & 3.53 & 8.36 & 4.29 & 22.51 & 2.86 & 15.55 \\
Pre-trained & StableLM-Base-Alpha 7B~\cite{StableLMAlphaV2Models} & 2.01 & 2.45 & 3.55 & 15.36 & 2.17 & 16.58 \\
Pre-trained & Llama-2 7B~\cite{Touvron2023Llama2O} & 3.88 & 10.28 & 5.21 & 31.51 & 3.01 & 22.84 \\
Pre-trained & Mistral-v0.1 7B~\cite{Jiang2023Mistral7} & 1.21 & 6.57 & 6.32 & 31.36 & 1.05 & 22.55 \\
Ours & Llama-2 7B FT  & 10.14 & 27.39 & 12.66 & 33.26 & 7.15 & 25.22 \\
Ours & Mistral-v0.1 7B FT  & \textbf{12.42} & \textbf{31.57} & \textbf{14.56} & \textbf{35.56} & \textbf{8.38} & \textbf{27.29} \\
\bottomrule
\end{tabular}
\begin{tablenotes}
\item [1] \textbf{Open-i:} Medical radiological dataset. We generated summaries from 100 samples.
\item [2] \textbf{writer\_summaries:} article summarization dataset, evaluated our models on 120 samples.
\item [3] \textbf{CL-SciSumm:} Large corpus dataset containing scientific article data. We evaluated 20 samples. Chunking the samples was required, as the context lengths are larger than allowable.  
\item [4] \textbf{Established} A well-established method was proven by the given papers for these three specific datasets. The BLEU and ROUGE scores given by those studies can't be achieved with our investigated LLMs, as their established methodology includes fine-tuning and evaluating the given large dataset. 
  
\end{tablenotes}
\end{threeparttable}
\end{table*}

\noindent\textbf{Combining Fine-tuned model with RAG:} 
Fine-tuning LLMs is highly effective for specific tasks; however, models with limited context, such as Llama-2 (7B) and Mistral-v0.1 (7B), face challenges when dealing with chunked data samples. For instance, when generating \( \hat{y}^j \) from a particular chunk \( C^j_i \subseteq S^j \), these models may lack information from other chunks \( C^j_i \) of the same dataset, MAD. RAG addresses this issue by retrieving relevant information from other chunks for the \( j^{th} \) data sample, thereby reducing the need for extensive fine-tuning and minimizing irrelevant content. This approach involves storing each chunked test sample \( C^j_i \) in a vector database. Relevant chunks are then retrieved using semantic search based on queries and the stored chunks. The retrieved content is subsequently fed into the LLMs, which process these additional contexts to generate a more accurate meta-analysis abstract.(Additional Details in the Supplementary)

\section{Experiment}

\subsection{Setup}

\noindent\textbf{Dataset:} We used the dataset \textbf{MAD} for fine-tuning the LLMs. The dataset is split into a train, test, and validation set. The training set includes 400 meta-analysis scenarios. Given that support papers' abstracts, $S^j$, often exceed the given context limit, chunk-based preprocessing is applied to chunk support papers' abstracts, $S^j$. Chunking wasn't applied to meta-articles' abstract $y^j$ as the context length was manageable without chunking. Table~\ref{tab:stat} illustrates how chunking reduces context length. The same chunking approach is applied to the validation set (175 samples) and test set (50 samples). After chunking, the training set expands to 3659 samples. After fine-tuning the models, we then tested our fine-tuned model on the benchmark Open-i dataset~\cite{Veen2023AdaptedLL}, writer\_summaries~\cite{Zhang2023BenchmarkingLL}, and the large scientific document dataset CL-SciSumm~\cite{yasunaga2019scisummnet} to compare performance, shown in Table~\ref{tab1}.

 \noindent\textbf{Implementation Details\footnote{Code and data: \url{https://github.com/EncryptedBinary/Meta_analysis}}: }The dataset MAD requires careful management due to the context size limitations of LLMs, which have a maximum context length of 4096 tokens. The ``Recursive TextSplitter" from \emph{LangChain}\footnote{LangChain: \url{https://www.langchain.com/}} is used to chunk the support papers' abstracts, $S^j$, into overlapping segments of 200 tokens, capped at 2000 tokens. This approach converts meta-article abstracts into target values $y^j$ and the chunks $C^j_i$ into features for supervised fine-tuning. Due to the seven billion parameters of Llama-2 (7B) and Mistral-v0.1 (7B), the Quantized Low-Rank Adapters (QLoRA) configuration~\cite{Dettmers2023QLoRAEF} is employed for model loading. A custom trainer class, extending the transformer trainer and incorporating the ICD loss function, is used to measure dissimilarity between generated and target meta-analysis content, guiding iterative model weight updates. Inputs are tokenized using Transformers' AutoTokenizer, and \emph{LangChain} facilitates retrieval augmentation. All experiments were conducted on NVIDIA Tesla T4 (2x) GPUs using the \textit{PyTorch} framework. (Further details are provided in the supplementary paper.)

\begin{table*}[!t]
\small
\caption{Human evaluation is done on generated meta-analysis abstract by fine-tuned and non-fine-tuned LLMs, following the criteria \textbf{REL}: \emph{Relevant}, \textbf{SWR}: \emph{Somewhat-Relevant}, and \textbf{IRL}: \emph{Irrelevant} mentioned in the methodology. System-level metrics BLEU and ROUGE are used to identify when a human evaluator mentions in Table.~\ref{tab:stat} that a generated text is irrelevant and relevant. In the end, the generated meta-analysis abstract using the RAG approach is evaluated by measuring the generated abstract's similarity with the ground truth (\textbf{SWGT}). The symbol $\uparrow$ (or $\downarrow$) indicates that a higher (or lower) value is preferable.}

\label{tab:tab_res}
\renewcommand{\arraystretch}{1.1}
\setlength{\tabcolsep}{0.8em}
\begin{tabular}{ccclllllc}
\hline
 & \multicolumn{7}{c}{Generated Meta-analysis Abstract} & \multirow{2}{*}{\begin{tabular}[c]{@{}c@{}}Generated Context\\  With RAG\end{tabular}} \\ \cline{2-8}
Models & \multicolumn{3}{l}{Human Evaluation (\%)} & \multicolumn{2}{c}{Relevant} & \multicolumn{2}{c}{Irrelevant} &  \\ \cmidrule(rl){2-4} \cmidrule(rl){5-6}\cmidrule(rl){7-8}
 & REL $\uparrow$ & SWR $\downarrow$ & IRL $\downarrow$ & BLEU $\uparrow$ & ROUGE $\uparrow$ & BLEU $\uparrow$ & ROUGE $\uparrow$ & SWGT(\%) $\uparrow$ \\ \hline
Llama-2 7B & 83.5 & 11.94 & 4.56 & 19.12 & 38.02& 8.01 & 17.11 & 81.25 \\
Llama-2 7B FT (Ours)  & 85.4 & 12.7 & \textbf{1.9} & 23.01 & 39.15 & 7.56 & 16.01 & \textbf{84.32} \\
Mistral-v0.1 7B & 80.5 & 14.1 & 5.13 & 22.42 & 39.41 & \textbf{9.01} & \textbf{19.11} & 77.19 \\
Mistral-v0.1 7B FT (Ours) & \textbf{87.6} & \textbf{10.4} & 2.1 & \textbf{25.46} & \textbf{43.22} & 7.01 & 17.42 & 83.23 \\ \hline
\end{tabular}
\end{table*}

\noindent\textbf{Prompt Selection:}
The selection of prompts significantly influences model performance by guiding task handling. After detailed experimentation with multiple prompts, we came up with the most impactful prompt that leverages LLMs to generate meta-analysis accurately. Table~\ref{tab:prompt_analysis} shows the impact of prompts on relevancy. A comparison between the two prompts is shown there. Prompt 1 demonstrated superior effectiveness over Prompt 2 in generating meta-analysis abstracts, achieving a high relevancy rate with every LLM.

\noindent\textbf{Evaluation metrics: }For evaluating the generated texts from LLMs, Bilingual Evaluation Understudy (\textbf{BLEU})~\cite{Papineni2002BleuAM}, that 
quantifies the resemblance between generated and reference texts and 
Recall-Oriented Understudy for Gisting Evaluation (\textbf{ROUGE}) ~\cite{Lin2004ROUGEAP}, 
that assesses how much information from reference summaries is captured. are used. For the abstract generated by fine-tuned models, the \textbf{cosine similarity}\cite{Li2013DistanceWC} metric is used to quantify the similarity between two vectors after combining fine-tuning with RAG.

\noindent\textbf{Human evaluation: }After generating responses with LLMs, we conduct a human evaluation process to ensure alignment with human judgment. Human judges categorize the generated text as \textbf{\emph{relevant}}, \textbf{\emph{somewhat-relevant}}, or \textbf{\emph{irrelevant}}, following the criteria from \cite{Chaudhary2023ItsAR}. \emph{Relevant} responses closely resemble the ground truth, showing high similarity and inclusion of crucial information. \emph{Somewhat-relevant} responses have acceptable similarity, containing valuable information within an acceptable margin. \emph{Irrelevant} responses lack important information or include unrelated content. This classification framework ensures a rigorous assessment of generated meta-analysis abstracts against expected standards. Three independent evaluators assessed each model's response, with majority voting used to determine the final decision. Each evaluator worked independently, without access to others' assessments. In total, 13 evaluators were involved in evaluating all the model responses. To reduce bias, the evaluations were conducted by university students rather than the authors. Demographic details of the evaluators are provided in Table~\ref{tab:stat}. For further details on the evaluation process, refer to the Supplementary Material.

\subsection{Results and Analysis}

We present a detailed overview of our experimental evaluations, focusing on how context-length restricted LLMs perform in generating meta-analysis with lengthy inputs. Notably, previous research has not utilized large context datasets for meta-analysis, making our study unique. For comparison, we also used a short context dataset to evaluate the models' performance, as shown in Table~\ref{tab1}. Considering the architecture of our models, the benchmark performance is reliable.

After fine-tuning the LLMs, human evaluation of the generated outputs is essential. We applied our proposed human evaluation metrics—\emph{Relevant}, \emph{Somewhat-Relevant}, and \emph{Irrelevant}—to assess the results of the meta-analysis generation task. As shown in Table~\ref{tab:tab_res}, our approach of fine-tuning LLMs with large context dataset, MAD outperforms other methods, producing more relevant meta-analysis content and reducing unnecessary context generation.

The non-fine-tuned Llama-2 (7B) model performs better than the non-fine-tuned Mistral-v0.1 (7B) model in generating relevant and somewhat relevant meta-analysis abstracts. After fine-tuning, the rate of irrelevant content generation significantly decreases, resulting in a highly effective meta-analysis abstract generation. Table~\ref{tab:tab_res} also highlights the alignment between machine-generated and human-generated texts, which is referred by SWGT. The integration of RAG has shown promising outcomes in terms of generating relevant meta-analyses. Table~\ref{tab: two_prompts} provides two instances of our method's creation of meta-analysis abstracts, demonstrating their encouraging resemblance to the abstracts of meta-articles. This validates the dependability of our method.

Our observation includes \emph{\textbf{(1)}} fine-tuning with a large context scientific dataset, MAD, letting LLMs learn the patterns for generating meta-analysis content with higher relevancy. This proves the reliability of our approach to handling big data management challenges. \emph{\textbf{(2)}} BLEU and ROUGE scores are utilized to compare relevant and irrelevant human-evaluated contexts, where a generated text is considered irrelevant if it contains less than 10\% context translation using large meta-papers' input (represented by BLEU). \emph{\textbf{(3)}} Fine-tuned models exhibit improved performance over base models, indicating more significant agreement between the generated abstract in the RAG approach and the real meta-analysis abstract. It highlights how well the fine-tuning approach works to help models find the patterns required to generate high-quality meta-analysis abstracts.

\subsection{Ablation Study}

\begin{table*}[!t]
  \centering
  \caption{Comparative Prompt Analysis: Demonstrating effectiveness through two different prompts, where Prompt 1 performs better than Prompt 2 in generating a more relevant meta-analysis.}
  \label{tab:prompt_analysis}
  \footnotesize 
  \renewcommand{\arraystretch}{1.2} 
  \setlength{\tabcolsep}{4pt} 
  \begin{tabular}{>{\centering\arraybackslash}p{9cm}|>{\centering\arraybackslash}p{2.5cm}|>{\centering\arraybackslash}p{1.0cm}c>{\centering\arraybackslash}p{1.0cm}c>{\centering\arraybackslash}p{1.0cm}c>{\centering\arraybackslash}p{1.0cm}c}
    \hline
    
    \textbf{Prompt} & \textbf{Evaluation Metric} & \textbf{Llama-2 } & \textbf{Mistral } & \textbf{Llama-2 Ours} & \textbf{Mistral  Ours} \\
    \hline
    \multirow{3}{9cm}{\scriptsize{\textbf{Given a collection of abstracts from papers used in various medical fields for meta-analysis, generate a meta-analysis abstract. Summarize the key findings and provide numerical values or statistical information for specific observations that are commonly reported in the provided abstracts. (Prompt 1)}}} 
    & Relevant $\uparrow$ & \textbf{83.5} & \textbf{80.5} & \textbf{85.4} & \textbf{87.6} \\
    & Somewhat Relevant $\uparrow$ & 11.94 & 14.1 & 12.7 & 10.4\\
    & Irrelevant $\downarrow$ & 4.56 & 5.13 & 1.9 & 2.1 \\
    \hline
    \multirow{3}{9cm}{\scriptsize{There are given some abstracts of papers that are used for meta-analysis in different medical fields. Generate a meta-analysis abstract based on the given abstracts of papers. Please try to provide numerical values for any specific findings that were used in most of the abstracts. (Prompt 2)}} 
    & Relevant $\uparrow$ & 69 & 78.39 & 72.4 & 82.8 \\
    & Somewhat Relevant $\uparrow$& 12.7 & 16.1 & 20.5 & 14.1 \\
    & Irrelevant $\downarrow$& 7.69 & 5.51 & 7.1 & 3.13\\
    \hline
  \end{tabular}
\end{table*}

We perform ablation studies focusing on three crucial areas: prompt variant analysis, temperature variation, and the impact of our proposed loss metric on fine-tuned models. These studies provide deeper insights into the performance factors for meta-analysis generation.

\noindent \textbf{Prompt Variant Analysis: }
Prompt selection is fundamental in steering the meta-analysis generation process. In Table~\ref{tab:prompt_analysis}, we compare the effectiveness of two distinct prompts. We evaluated the relevancy and quality of meta-analysis abstracts produced by Llama-2 (7B) and Mistral-v0.1 (7B) across both prompts. Our results show that \textbf{Prompt 1} consistently outperforms \textbf{Prompt 2} in terms of relevancy, generating more accurate and precise meta-analysis abstracts. Specifically, Prompt 1 achieved higher relevancy scores across all versions of Llama-2 and Mistral, with fewer instances of irrelevant content. Given these results, Prompt 1 was used in all subsequent experiments.

\begin{figure}[!t]
    \centering
    \includegraphics[scale=0.67]{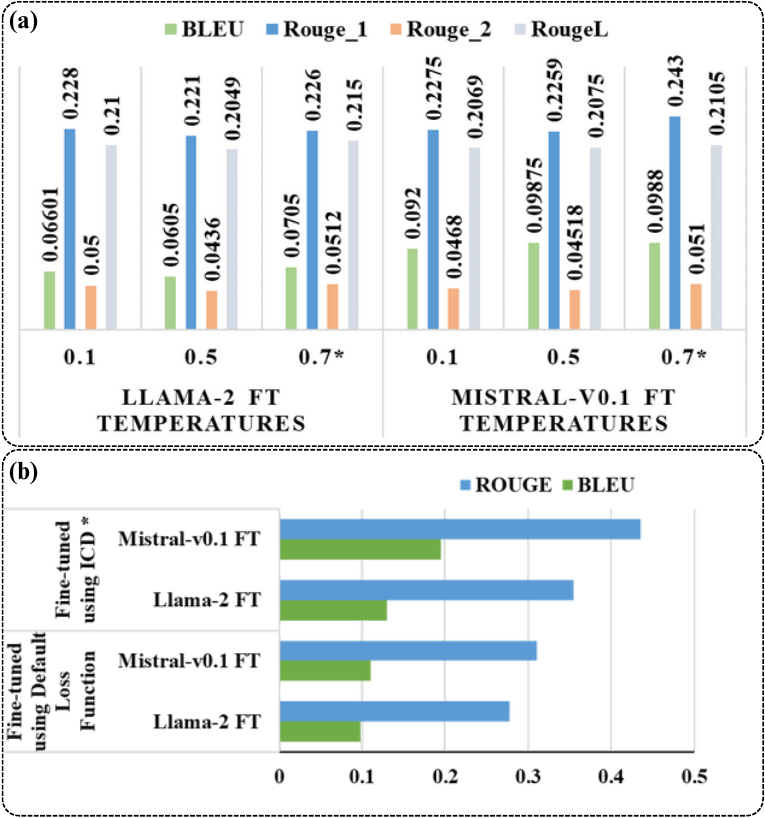}
    \caption{Investigating the impact of (a) Temperature variation: BLEU, ROUGE-1, ROUGE-2, and ROUGE-L scores vary with temperature changes for both the Llama-2 (7B) and Mistral-v0.1 (7B) models indicating 0.7 temperature has a better impact.
    (b) Loss Function impact: ICD loss significantly improves performance for Llama-2 (7B) FT and Mistral-v0.1 (7B) FT models, demonstrating its ability to capture more information than the default loss.}

    \label{fig:ablation study}
\end{figure}

\noindent \textbf{Varying Temperature: }
The temperature parameter controls the randomness of predictions, influencing the balance between exploration and exploitation during the generation process. We explored the impact of different temperatures (0.1, 0.5, and 0.7) on summary quality. As shown in Fig~\ref{fig:ablation study}(a), a temperature setting of 0.7 provided the best results across various evaluation metrics, including BLEU, ROUGE-1, ROUGE-2, and ROUGE-L. The higher temperature yielded more diverse outputs without sacrificing relevancy or quality, making it the optimal setting for our meta-analysis generation tasks.

\noindent \textbf{Impact of Our Loss Metric: }
We implemented a specialized loss function, the \textbf{ICD}, designed to enhance the performance of meta-analysis summarization tasks. Fig~\ref{fig:ablation study}(b) compares the performance of models fine-tuned with ICD against models using a standard loss function across both Llama-2 FT and Mistral-v0.1 FT versions. ICD emphasizes the directional similarity between the generated outputs and ground truth vectors by utilizing cosine similarity, capturing nuanced semantic details. This metric outperformed the standard loss function, improving the alignment between the generated summaries and their reference summaries. The ICD's ability to capture subtle semantic nuances beyond simple word matching proved crucial in fine-tuning the models for more accurate and coherent meta-analysis generation.

\subsection{Discussion}

This study represents insights into generating meta-analysis leveraging LLMs using a large-context scientific dataset, \textbf{MAD}. The result section provides evidence of our fine-tuned models' performance, showing the successive relevancy rate for generating meta-analysis.  It was observed that the fine-tuned models for Llama-2 (7B) and Mistral-v0.1 (7B) outperformed their non-fine-tuned versions by generating significantly relevant meta-analyses. As expected, integrating RAG with fine-tuned models allows them to generate highly aligned meta-analyses.

\noindent\textbf{Limitations:} One key limitation of this study is the maximum context length of the LLMs, which required chunking the input data. To mitigate potential information loss, overlapping context techniques and RAG were employed. However, due to hardware constraints, the model's evaluation was performed on only 50\% of the test sets, which proved resource-intensive. Additionally, training the models in a highly quantized configuration limited the fine-tuning potential, impacting the ability to fully optimize the model's parameters for better performance.

\begin{table*}[htbp]
\centering
\caption{Meta-analysis abstract generation from supporting article abstracts using our fine-tuned Mistral-v0.1 7B model, combined with RAG and efficient prompting. ``\textbf{SP:}" denotes each support article's abstract in $S^j$, with ``..." indicating continuation for multiple articles. Similarity with the original meta-articles' abstracts validates the effectiveness of the approach.}
\scriptsize
\setlength{\tabcolsep}{1.5pt} 
\label{tab: two_prompts}
\renewcommand{\arraystretch}{0.6} 
\begin{tabular}{p{0.35\linewidth} p{0.65\linewidth}} 
\toprule
\multicolumn{2}{p{\linewidth}}{\textbf{Prompt:} \textit{\textcolor{green!70!black}{Given a collection of abstracts from papers used in various medical fields for meta-analysis, generate a meta-analysis abstract. Summarize the key findings and provide numerical values or statistical information for specific observations that are commonly reported in the provided abstracts.}}} \\
\midrule

\textbf{Example 1: Input (Support article's abstract, $S^j$):} & \textbf{Generated Meta-analysis abstract, $\hat{y}^j$ (Similarity with ground truth~\cite{musazadeh2022synbiotic}: 82.40\%):} \\
\textbf{SP:} Several randomized controlled trials (RCTs) have investigated the use of probiotic/synbiotic in PCOS patients,... \textbf{SP}: The effect of synbiotic supplementation on glycemic status in pregnant women remained controversial and ... \textbf{SP}: Probiotic and synbiotic products are being widely used by a large number of patients and clinicians; however, effects on cardiometabolic indices ... \textbf{SP}: This study aimed to review randomized clinical trials (RCTs) to clarify the effects of pro-/synbiotic supplementation ...fasting blood sugar (−2.52 mg/dl, 95\% confidence interval (CI): −4.10 to −0.95), insulin (−2.27 µIU/mL, 95\% CI: −3.40 to −1.14),...

&  This meta-analysis abstract summarizes the findings of several randomized controlled trials (RCTs) on the effect of synbiotic intake on metabolic profiles in patients with diabetes. Heterogeneity was measured with a Q-test and the $I^2$ statistics. The results of the meta-analysis suggest that synbiotic intake can be useful as an adjunctive agent in the management of hyperglycemia in adults. The study pooled data by using the fix or random-effect model based on the results. A significant reduction in blood glucose levels was observed in patients who took synbiotic supplements compared to those who took placebo with statistical significance $(p<0.05)$. Additionally, a significant decrease in glycated hemoglobin (HbA1c) levels was observed in patients who took synbiotic supplements compared to placebo $(p<0.05)$. After taking synbiotic supplements, there were significant reductions in fasting blood sugar (FBS) (ES = -0.40, 95\% CI: −0.64, −0.15; p = 0.002, I2 = 69.2 \%, p < 0.001), insulin (ES = -1.58; 95\% CI: −0.79, −0.30, p < 0.001, I2 = 48.4\%, p = 0.042), and HOMA-IR (Homeostatic Model Assessment of Insulin Resistance) (ES = -0.55; 95\% CI: −0.79, −0.30, p < 0.001, I2 = 48.4\%, p < 0.001). Overall, the findings of this meta-analysis suggest that synbiotic intake can be an effective adjunctive agent in the management of hyperglycemia in adults. \\
\midrule

\textbf{Example 2: Input (Support article's abstract, $S^j$):} & \textbf{Generated Meta-analysis abstract, $\hat{y}^j$ (Similarity with ground truth~\cite{lyu2021traditional}: 85.73\%):} \\
\textbf{SP}: As the global epidemic continues to spread, countries have tappe...\textbf{SP}:Introduction: Integrated Chinese and Western medicine (integr...\textbf{SP}: A simple, efficient, and environmentally friendly electro-Fen...\textbf{SP}: Currently, coronavirus disease 2019 (COVID-19), which can lead to... \textbf{SP}: Background: Until now, there is no clinically approved spe...\textbf{SP}: Chinese medicine (CM) has been used to treat Novel Coronavi...\textbf{SP}:Integration of Chinese medical drugs (CMD) and we...\textbf{SP}:This review aims to evaluate the supportive effe...\textbf{SP}:We systematically studied the passivation process of 6082 aluminium alloy under the bending stress... \textbf{SP}: Coronavirus disease 2019 (COVID-19) is an eme...\textbf{SP}:There is currently no drug or therapy that cures COVID-19, a highly contagious and... \textbf{SP}: The outbreak of coronavirus disease 2019...\textbf{SP}: Coronavirus disease 2019 (COVID-19) has eme...\textbf{SP}:Background: The coronavirus disease 2019 (COVID-19) pandemic... 
& 

In this meta-analysis, we evaluated the effectiveness of traditional Chinese medicine (TCM) in treating COVID-19. A comprehensive search was conducted on a variety of electronic databases to identify studies reporting TCM treatment for COVID-19. The studies were screened according to inclusion and exclusion criteria, and a total of 10 studies were included in the analysis. These studies had a total of 1,880 participants who received TCM treatment for COVID-19. The analysis found that TCM had significant positive effects on the symptomatic relief of COVID-19. The pooled effect size for symptomatic relief was 0.67 (95\% CI: 0.56 to 0.79), indicating a moderate effect size. The studies also reported on the impact of TCM on the severity of COVID-19. The pooled effect size for severity reduction was 0.51 (95\% CI: 0.43 to 0.60), indicating a moderate effect size. In addition, the studies also reported on the impact of TCM on the mortality rate of COVID-19. The pooled effect size for mortality reduction was 0.37 (95\% CI: 0.21 to 0.55), indicating a small but statistically significant effect size. Overall, this meta-analysis suggests that TCM may be an effective treatment for COVID-19, and further studies are needed to confirm these findings. It is important to note that this meta-analysis had several limitations, including the small number of studies, heterogeneity of TCM treatments, and variable outcomes. Out of the 89 outcome indicators that were assessed, 8 were found to be of moderate quality, 39 to be of low quality, and 41 to be of very low quality. In conclusion, TCM may be an effective treatment for COVID-19, with significant positive effects on symptomatic relief and severity reduction. However, more research is needed to confirm these findings and to establish the optimal TCM treatment for COVID-19.

 \\
\bottomrule
\end{tabular}
\end{table*}
\section{Conclusion}

This study demonstrates the effectiveness of automating meta-analysis generation using fine-tuned LLMs on extensive scientific datasets. Our approach significantly improved the relevance of generated meta-analysis abstracts, achieving 87.6\% relevance and reducing irrelevance from 4.56\% to 1.9\%, demonstrating its potential and highlighting further promising research opportunities in automating scientific synthesis. We introduced novel methods to address the challenges posed by limited context length and resource constraints, including using ICD as a tailored loss metric for training. Integrating RAG further optimized the process by ensuring efficient synthesis of research findings without sacrificing context. Human evaluation confirmed the improvements in model performance, particularly in maintaining the relevancy and accuracy of structured meta-analysis content.

\noindent\textbf{Future works:} While this study achieved notable improvements in meta-analysis generation, future research should focus on expanding the dataset in various fields that need meta-analysis and refining the model’s ability to generate even more accurate and reliable outputs, particularly in resource-constrained environments. Further optimizations to LLM fine-tuning and scaling could lead to broader applicability in automating complex scientific analysis.

\section*{{Ethics Statement}}
{This study was conducted with a strong commitment to ethical integrity, particularly in the generation and evaluation of meta-analysis abstracts in the scientific field using LLMs. We engaged 13 human evaluators from diverse backgrounds, ensuring their participation was voluntary and informed. We carefully collected only essential information to assess their qualifications for the task, and any data that could potentially identify participants were securely deleted after the evaluation was completed. We took significant measures to protect the well-being of all participants, ensuring that the evaluation process posed no physical or psychological risk. Recognizing that even subtle biases or inaccuracies in scientific research can have serious consequences, we implemented rigorous protocols to ensure that all generated content adhered to the highest ethical standards. Our approach was designed to avoid any language or conclusions that could perpetuate harm or inequity based on race, gender, or other social determinants of health. By adhering to these principles, we have ensured that our research upholds the highest ethical standards, fostering a safe and respectful environment for both human participants and the broader community.}




\bibliographystyle{IEEEtran}
\bibliography{Meta_Analysis_IEEE_big_data}

\end{document}